\documentclass{article}

\usepackage[dblblindworkshop, final]{neurips_2025}

\usepackage[utf8]{inputenc} 
\usepackage[T1]{fontenc}    
\usepackage{hyperref}       
\usepackage{url}            
\usepackage{booktabs}       
\usepackage{amsfonts}       
\usepackage{nicefrac}       
\usepackage{microtype}      
\usepackage{xcolor}  
\definecolor{darkblue}{rgb}{0,0,.4}
\usepackage{hyperref}
\usepackage{tabularx}
\usepackage{array}

\hypersetup{
    colorlinks=true,
    linkcolor=darkblue,
    citecolor=darkblue,
    filecolor=magenta,      
    urlcolor=cyan
}

\title{Securing Dual-Use Pathogen Data of Concern}
\workshoptitle{Biosecurity Safeguards for Generative AI}

\author{%
Doni Bloomfield\thanks{These authors contributed equally.}\\
Fordham University School of Law\\
\texttt{doni.bloomfield@fordham.edu}
\And
Allison Berke\footnotemark[1]\\
RAND
\And
Moritz S.\ Hanke\\
Center for Health Security\\
Johns Hopkins Bloomberg School of Public Health
\And
Aaron Maiwald\\
Department of Statistics\\
University of Oxford
\And
James R.\ M.\ Black\\
Center for Health Security\\
Johns Hopkins Bloomberg School of Public Health
\And
Toby Webster\\
RAND Europe
\And
Tina Hernandez\mbox{-}Boussard\\
Stanford University School of Medicine
\And
Oliver M. Crook\\
Department of Chemistry \& Kavli Institute for Nanoscience Discovery\\
University of Oxford
\And
Jassi Pannu\footnotemark[1]\\
Center for Health Security\\
Johns Hopkins Bloomberg School of Public Health\\
\texttt{pannu@jhu.edu}}

\begin{document}

\maketitle

\begin{abstract}
Training data is an essential input into creating competent artificial intelligence (AI) models.  AI models for biology are trained on large volumes of data, including data related to biological sequences, structures, images, and functions. The type of data used to train a model is intimately tied to the capabilities it ultimately possesses--including those of biosecurity concern. For this reason, an international group of more than 100 researchers at the recent 50th anniversary Asilomar Conference endorsed data controls to prevent the use of AI for harmful applications such as bioweapons development. To help design such controls, we introduce a five-tier Biosecurity Data Level (BDL) framework for categorizing pathogen data. Each level contains specific data types, based on their expected ability to contribute to capabilities of concern when used to train AI models. For each BDL tier, we propose technical restrictions appropriate to its level of risk. Finally, we outline a novel governance framework for newly created dual-use pathogen data. In a world with widely accessible computational and coding resources, data controls may be among the most high-leverage interventions available to reduce the proliferation of concerning biological AI capabilities.
\end{abstract}

\section{A Taxonomy of Dual-Use Pathogen Data of Concern}

Open access to biological data accelerates scientific discovery, fosters collaboration, and promotes reproducibility in research. Given these benefits, and the importance of free expression, governments should be wary of imposing barriers to data access. 

At the same time, we argue that the default of open access is not always appropriate for the narrow subset of newly created biological data that could confer concerning capabilities on AI models \citep{bromberg20253}. Recent evidence indicates that viral sequence data is one such subset. Excluding viral proteins from the training data of the generative protein model ESM3 noticeably reduced its performance on benchmarks related to viruses \citep{hayes_simulating_2025}. Likewise, excluding genomes of viruses infecting eukaryotes from the training data of the genomic language model Evo 2 led to poor performance on fitness-prediction tasks related to the genomes of viruses infecting humans \citep{brixi_genome_2025}.  Work by a subset of this paper's authors shows that fine-tuning Evo 2 on sequences from human-infecting viruses (originally excluded from the Evo 2 training set) reduced the model's perplexity in predicting withheld human-infecting viral sequences (i.e., those not in the fine-tuning dataset) compared with the pretrained model \citep{black_openweight_genome_lm_safeguards}. The fine-tuned model was also better at predicting SARS-CoV-2 immune escape variants than the pretrained model. Taken together, this early empirical evidence bolsters the theoretically plausible view that data on human-infecting pathogens is an important ingredient in creating AI models that can design pathogens with enhanced pandemic potential.

In this paper, we present a framework for regulating access to the most concerning pathogen data. The framework is designed to reduce misuse risks while minimizing costs to research and open-science norms. We begin by summarizing biological AI model capabilities of concern (§ 1.1) before turning to our framework for categorizing pathogen data into distinct risk tiers (§ 1.2). This schema categorizes data based on its expected contribution to capabilities of concern, and we view this preliminary work as a vocabulary and set of hypotheses for further empirical validation (§ 1.3). We then introduce a set of technical data-access controls (§§ 2.1, 2.2) that can be applied with escalating stringency as a data type's risk increases (Table 1). Finally, we close with a proposed governance system for turning a version of the framework into effective, balanced regulations (§ 3).

Our focus is on defining and governing the most concerning datasets before they are generally available to AI developers. This focus is relatively rare in the AI governance literature, particularly that related to safety, which more typically develops approaches to governing model training (e.g., data exclusion), access to model weights (e.g., cyber protections for model weights), or model outputs (e.g., safe refusals)  \citep{Anthropic2025PretrainingFiltering} \citep{CISA2025AIDataSecurity} \citep{OpenAI2025PreparednessV2}   \citep{RAND2024SecuringWeights}; but see \citep{Berke2025AIEnabledBioDesign}. We draw inspiration from the literature on securing private data, where upstream data governance is more common  \citep{OECD2025TrustworthyModelsPETs} \citep{Voisin2021Passports}.

\subsection{Biological AI Capabilities of Concern}

The stakes of biological data governance are high, as AI models could help create severe biological threats. AI models may contribute to particularly consequential biological risks, for example by enabling, accelerating, or simplifying "the creation of transmissible biological constructs that could lead to human pandemics, or similar pandemic-like events in animals, plants or the environment” \citep{pannu_dual-use_2025}. Researchers have proposed several AI model “capabilities of concern” that may enable such high-consequence harms, for example, a model's ability to predict or generate more transmissible variants for pandemic pathogens \citep{pannu_dual-use_2025}. These capabilities of concern resemble the capabilities that scientists and policymakers have homed in on as most concerning after decades of debate about dual-use biological research. 

\subsection{Biosecurity Data Levels (BDLs)}

To analyze datasets that pose a substantial risk of increasing capabilities of concern, we introduce a five-tier Biosecurity Data Level (BDL) framework for categorizing pathogen data. While the naming of these tiers references the Biosecurity Level (BSL) system, there is no other relationship between the two systems, and data at a particular BDL level does not correspond to experiments performed at a particular BSL level. The BDL tiers range from BDL-0 to BDL-4. At each level, we propose securing and overseeing data with progressively higher risk-benefit ratios. We outline specific data types for BDL-0 to BDL-4, based on their expected ability to contribute to capabilities of concern with continuously increasing pandemic risk. In Section 2, we link each BDL tier to layered technical restrictions appropriate to its level of risk. BDL-0, which encompasses the vast majority of biological data, should be subject to no restrictions. We caution that our example data types are preliminary, and that more empirical investigation is needed to assess the capability contribution of different data types. 

\textbf{BDL-0:} All data types other than those named below. 

\textbf{BDL-1:} Data reasonably anticipated to enable an AI system to learn general patterns of eukaryote-infecting viral genome and protein composition.  Corresponding BDL-1 data types may include: 
\begin{itemize}
\item Eukaryote-infecting viral sequencing data (genomic or protein level) \citep{brixi_genome_2025}
\item Eukaryote-infecting viral protein structure data (experimentally generated, or generated from sequence via structure prediction models) \citep{hayes_simulating_2025}
\end{itemize}

\textbf{BDL-2:} Data reasonably anticipated to enable an AI system to learn the following properties of viruses from families likely capable of causing pandemics in non-human animals: zoonotic crossover, environmental stability, host range, susceptibility of host populations, and evasion of diagnostics or detection. Corresponding BDL-2 data types may include: 
\begin{itemize}
\item Sequencing data with functional annotations for viral families capable of causing pandemics in non-human animals.
\item Functional assays from relevant viral families regarding features conveying species host range and tropism (e.g., binding affinity for various human and animal tissues).
\item Protein-protein interaction data for relevant receptors and proteins for zoonotic crossover, host infectivity, and tropism, including binding affinity.
\item Diagnostic data comparing wild-type and viral variants.
\item Data on environmental stability for relevant viral proteins.
\end{itemize}

\textbf{BDL-3:} Data reasonably anticipated to enable an AI system to learn the following properties of viruses from families likely to infect humans: transmissibility, virulence, immune evasion, and resistance to medical countermeasures.  Corresponding BDL-3 data types may include: 
\begin{itemize}
\item Sequencing data with functional annotations from viral families demonstrably or likely pathogenic to humans, including pathogens that infect humanized animal models or non-human animals that are known candidates for causing zoonotic crossover (e.g., primates, fruit bats, and mustelids) 
\item Functional assays from relevant viral families regarding features conveying transmissibility, virulence, immune evasion, or resistance to medical countermeasures.
\item Protein-protein (or other modality) interaction data from viral families demonstrably or likely pathogenic to humans for relevant proteins for transmissibility, virulence, immune evasion or medical countermeasure resistance.
\item Phenotypic data from animal models, as well as transmission rate data linked to viral variants.
\end{itemize}

\textbf{BDL-4:} Data reasonably anticipated to enable an AI system to learn the following properties for viral variants from families likely to cause pandemics in humans: enhanced transmissibility, enhanced virulence, or enhanced immune evasion compared to known wild-types. Corresponding BDL-4 data types may include: 
\begin{itemize}
\item Sequencing data with functional annotations from viral variants from families with pandemic potential in humans demonstrated to enhance transmissibility, virulence, or immune evasion.
\item Functional assays from viral families with pandemic potential in humans regarding features conveying transmissibility, virulence, and immune evasion. 
\item Protein-protein interaction data from viral families with pandemic potential in humans for relevant receptors and proteins for transmissibility, virulence, and immune evasion (e.g., antibody binding affinity).
\end{itemize}

\subsection{Taxonomy Limitations and Next Steps}

The presented taxonomy assumes that data enables AI capabilities closely related to the information that data encapsulates. For instance, we assume that sequence data from human-infecting viruses is required to train a model with the capability of accurately predicting or generating functional (or at least realistic) sequences for human-infecting viruses. Most importantly, this taxonomy assumes that functional data for viruses, e.g., deep mutational scanning or virus-host protein-protein interaction data, is particularly relevant for enabling capabilities of concern. This assumption is motivated by the view that complexity and noise at the cellular and organismal level render it difficult to predict the relationship between pathogen genomes and functional outcomes (such as transmissibility and virulence) on the basis of genetic information alone.

However, several models predicting viral fitness simply rely on sequencing data, not additional functional data (e.g., \citep{king_generative_2025}).  More broadly, if biological foundation models can sufficiently generalize using large volumes of limited data types (such as sequence data), our proposed BDL framework would require revision. Further empirical assessment is needed to determine the degree to which specific biological data types are particularly effective at enabling corresponding biological AI capabilities of concern. Systematic evaluation of the effect of inclusion or exclusion of particular datasets on model capabilities would be helpful in this regard. As we discuss in Section 3, policymakers and scientists should use such empirical work to create technical documents translating these high-level categories into objective, readily applicable classifications.  

We welcome further discussion and refinement of our initial taxonomy, the included data types, and their assigned BDLs. This should ultimately allow for a comprehensive taxonomy of dual-use pathogen data of concern that can be leveraged for responsible oversight by researchers, funders, and governments.

Similar to rules associated with laboratory biosafety levels (BSLs), BDLs are designed to give legitimate researchers regulated access to pathogen data to pursue beneficial public health and pandemic preparedness research, while limiting risks of deliberate and accidental misuse through unrestricted access. Although the framework introduces limitations to typical procedures around open data sharing, a BDL oversight regime is likely preferable to potential governance alternatives that may arise in an ad-hoc, reactive manner. To ensure that policymakers and researchers can consistently apply the framework, and to reduce its cost, BDL definitions should be clear and narrowly scoped. In addition, policymakers should remove BDL restrictions on data regarding pathogens causing a significant and ongoing disease outbreak, as real-time data sharing in these events can yield large public health benefits. 

Laws harnessing the BDL framework may also bolster open-source model development and reduce the degree of complexity required for built-in technical safeguards for some models. If researchers adequately control data required for adversarial fine-tuning of models--for instance, functionally annotated data for human-infecting viruses--that lessens concerns about releasing models openly. In addition, scientists and policymakers have argued that the absence of data standardization and high-quality datasets has slowed biological model development. Increasing data standardization through BDLs could improve training data quality for legitimate and beneficial training and fine-tuning applications.

Although categorizing dual-use data will likely prove important in the quest to reduce AI risk, it will be of greatest use in combination with competent technical tools and sensible data regulations. To pair our taxonomy with the institutions needed to implement it, in the following two sections we briefly outline a suite of relevant tools and propose an initial governance framework. 

\section{Technical approaches for oversight of dual-use pathogen data}

Policymakers and researchers will need to develop innovative technical approaches to data security to facilitate adequate oversight while minimizing impact on legitimate researchers. This section first outlines technical tools that are largely designed to govern data that can be exported to researchers. The goal with such tools is to ensure that data providers can confirm that only appropriate users are accessing the data, and, if necessary, to track whether data is being or has been misused. Next, we discuss how Trusted Research Environments can allow legitimate scientists to use especially sensitive data without directly accessing or exporting the data. In Table 1, we propose tools that should be required for each BDL tier. 

\subsection{Novel cybersecurity-informed technical methods}

\emph{Watermarking/fingerprinting} involves embedding hidden, unique identifiers in datasets, sequences, metadata or models, e.g., using synonymous codon substitutions \citep{jupiter_dna_2010, zhang_foldmark_2024}. This approach would help officials to trace leaked or misused data back to its source. Currently, maintaining watermark integrity through edits, synthesis or lab work is technically challenging. Improvement in methods should be focused on ensuring watermarks cannot be erased, and are hard to plant as a false flag. Finally, watermarking must not reduce data quality. 

\emph{Data provenance and audit logs} allow for recording of who accessed or changed data, when, and how, with cryptographic signatures to prevent tampering. In the biodata context, such an approach would enable post-event attribution, compliance checks, and chain-of-custody tracking for sensitive biological datasets and gene synthesis requests. Successful audit logging and tracking of data provenance requires consistent integration across all data tools and storage, which is often resisted due to perceived administrative overhead. 

\emph{Anomaly detection} is the statistical or machine-learning based monitoring of user behavior, such as access frequency, or the use of unusual dataset combinations. For example, this could include modeling of suspicious downloads one week before a user exits their institution of employment, or noting the sudden surge in volume of data downloaded by a user. Providers could monitor geographical anomalies, such as when a user accesses data from an unexpected country. Together, these technical approaches enable flagging suspicious activity early in a privacy preserving manner, as reading the content is not required. Anomaly detection may be a useful approach to spotting misuse of sensitive genome or sequence data, data theft, or hacked user accounts. Limitations currently include the lack of good activity baselines to avoid false positives; these baselines could be obtained through metadata sharing or red teaming. 

\emph{Honeytokens and decoys} involve planting synthetic "fake" data or sequences that are unlikely for certain legitimate workflows. Ideally, decoys would be designed to be context-aware, such that they mimic realistic research datasets but contain subtle biological impossibilities, such as a protein structure that would not fold or express correctly. Decoys are commonly used in cybersecurity to detect security breaches, as once a decoy is accessed, it can immediately signal a potential misuse attempt. However, decoys need to be carefully designed so they are not accidentally used in beneficial research or by legitimate researchers. 

\emph{Federated learning} permits training anomaly detection or risk models across multiple institutions without centralizing raw data. This makes it easier for authorities to detect misuse by learning from distributed, rare-event patterns while keeping sensitive data local. However, implementing federated learning across organizations requires cross-institutional trust, compatible infrastructure, and shared agreement on oversight policies. 

\emph{Confidential computing} is a technique that uses hardware-protected enclaves or secure virtual machines (VMs) to run sensitive computations. It requires specialized hardware that is not currently broadly accessible to academic researchers and is associated with increased cost and integration complexity. Confidential computing prevents unauthorized parties (even system administrators) from seeing raw sensitive data during analysis or synthesis screening. Applying this technique to durably mask sensitive data during research processes is an unsolved problem and may be very difficult or impossible.

\emph{Automated risk scoring} enables risk assessment at scale and in real time, using automated methods to score the potential risk of a given workflow before its execution. These risk assessments utilize factors such as sequence similarity to controlled items, provenance, and user trust scores. An elevated risk score triggers action, stopping high-risk workflows before they happen or flagging them for human review (as is currently done for sensitive gene synthesis order requests). For successful automated risk scoring, accurate risk criteria and low false-positive rates are needed, otherwise this method may unduly block legitimate work. 

\emph{Behavioral biometrics} leverage the unique patterns in how individual users interact with their computers, such as how they type and move their cursor. These patterns create a digital fingerprint of a user that is challenging to fake. However, in order to implement this method, months of usage and pattern tracking for each user are needed. 

\emph{Hardware keys for data access} are familiar across the cybersecurity landscape. In the biology context, this would involve multi-factor authentication in order to access sensitive biological data. 

\emph{API rate limiting} allows data providers to place limits on the quantity of data a user can access over a given time period. Rate limiting avoids bulk data exfiltration or theft, in cases where bulk data downloading is not required for most research purposes. 

\subsection{Trusted research environments}

Trusted research environments (TREs) offer a practical way to oversee biological data that is valuable for training AI without making the most hazardous content broadly available. Pairing narrow access restrictions for the highest BDL tiers with government investment in standardized TREs would allow qualified researchers to do high-value work under strong safeguards, including rigorous user credentialing and continuous usage monitoring. In this section, we explain how TREs should be used to secure sensitive biological data for AI training.

Operationally, TREs use a model in which approved users bring code to secured data rather than exporting records, and only risk-checked outputs can leave. In the field of health research, some examples include OpenSAFELY in England, the SAIL Databank’s UK Secure Research Platform, the commercial health data platform DNAnexus, and the National COVID Cohort Collaborative (N3C) enclave in the United States \citep{nab2024opensafely,haendel2021national,green2023present}. These examples demonstrate that this approach scales while preserving analytical utility, albeit outside the domain of large-scale AI training runs.  TREs operate under principles of tiered data identifiability, strict ingress/egress controls, two-factor authentication, containerized/reproducible tooling, and independent output checking, typically organized under the Five Safes framework (Safe People, Projects, Data, Settings, Outputs) \citep{nab2024opensafely}. These practices map cleanly onto biosecurity-motivated governance for biological training data \citep{bloomfield2024ai}. 

\emph{Requirements for TREs that steward biological data used to train AI models}: \newline
1. Risk-tiered data classification and minimization (Safe Data): Biological inputs are classified by biosecurity hazard, such as the above proposed BDLs, and stricter usage controls are assigned to higher-risk tiers. By default, the TRE stores only the variables necessary for the stated aims of the researcher or analysis, and prefers to store derived features (e.g., k-mer summaries, embeddings) over raw sequences when feasible. TREs such as SAIL and OpenSAFELY implement tiering and pseudonymization measures that retain utility while limiting re-identification and sensitive content exposure  \citep{nab2024opensafely}. 

2. Locked-down compute (Safe Settings): Implementation of a TRE requires isolated VMs/containers, no inbound internet, controlled package installation, and secure GPU access when model training is needed. Some TREs also require reproducible workflows, keeping a public log of every job run against real data. Where model training occurs inside the TRE, the system should implement measures to prevent training-data extraction and bar the export of raw model weights for models trained on high-hazard content \citep{nab2024opensafely,abadi2016deep,carlini2019secret}. Implementing TREs for large-scale model training runs may require special compute clusters or partnerships with hyperscalers capable of providing sufficient capacity.

3. Strong identity, training, and contracting (Safe People \& Projects): TRE implementation enforces multi-factor authentication, approving researchers and institutions via Data Access Agreements that bind use to public-interest projects and prohibit malicious or unduly dangerous aims. Some also require biosecurity and disclosure-control training \citep{brophy2023towards}. TREs can use automated risk scoring, discussed above, to help guard against assisting dangerous projects. 

4. Egress control with both privacy and biosecurity checks (Safe Outputs): Some TREs use human-in-the-loop “output checking”, which could be extended to include biosecurity screening to block release of dangerous biological sequences, executable protocols, or optimization guidance. This can be used to restrict output to aggregate measures, statistics, or vetted and non-actionable textual summaries of potentially dangerous protocols. The TRE could require that researchers provide data provenance information and a statement of any identified biosecurity concerns with the data being provided. The TRE could also provide model provenance information and model cards that specify prohibited uses of the TRE-secured data, safety evaluations that must be performed before model extraction, and deployment guardrails that model developers must agree to use \citep{gebru2021datasheets}.

\begin{table}
  \caption{Potential technical security correspondence per Biosecurity Data Level (BDL)}
  \label{table-1}
  \centering
  \begin{tabular}{ m{3cm} m{3cm} m{5cm} }
   \toprule
 \textbf{BDL} & \textbf{Access Controls} & \textbf{Technical Security Methods} \\
 \midrule
 BDL-0 & Optional monitoring & \\
 \hline
 BDL-1  & Managed access  & Provenance and audit logs, anomaly detection, API rate limiting, risk scoring, intent logging  \\
  \hline
 BDL-2  & Identity accreditation  & As above plus hardware keys, enhanced anomaly detection (cross-institutional), watermarking  \\
  \hline
 BDL-3  & Use approval  & As above with TRE integration including confidential computing, behavioral biometrics, sophisticated honeytokens, real-time risk scoring\\
  \hline
 BDL-4  & Pre-screening  & As above with TRE integration including provenance recording, facial recognition, and cross-institutional federated learning\\
  \bottomrule
  \end{tabular}
\end{table}

\section{Requirements for effective governance of dual-use pathogen data}

Reducing AI biosecurity risks requires technical solutions—but technical solutions do not exist in a vacuum. To effectively reduce the risks associated with dual-use biological data, we must combine technical advances with effective institutional design. Here, we propose a governance framework to oversee newly created (rather than preexisting) dual-use pathogen data of concern. Although our proposal is based on the U.S. institutional experience, it should have broader application internationally.

In governing dual-use biological data, we need not write on a blank slate. By regulating biological research over decades, the U.S. government has provided valuable lessons in what does and does not work. The National Institutes of Health (NIH), for example, has crafted rules to protect human clinical trial subjects, reduce the risk of infectious agents escaping laboratories, safeguard patient genetic data, and control the diffusion of experimental knowledge liable to misuse \citep{wolf2013behind,wright1994molecular,pannu_dual-use_2025}. With all of these rules, the government has sought to balance safety with scientific autonomy and flexibility, to varying degrees of success. In designing rules to regulate biological training data, we can build on past successes and learn from past mistakes. 

Our framework relies on four main principles derived from past research oversight efforts. 

First, data governance rules should be readily applicable, leaving little room for local, subjective decisions and ensuring that like cases are treated alike. The U.S. government has at times tasked non-experts overseeing research with applying vague and uncertain standards. For example, dual-use research of concern policies ask scientists whether it is “reasonably anticipated” that a given experiment will result in a more infectious pathogen. Similarly, IRB rules require committee members to regularly opine about the net costs and benefits of studies far outside their areas of expertise \citep{gunsalus2006mission,schneider2015censor}. Given the technical and relatively narrow nature of pathogen data risks, we should design more reliable and objective rules in this domain. Technically clear policies will have the additional advantage of reducing the amount of time that scientists spend debating what rules apply to their research. 

Second, data oversight rules should be enforced by a neutral arbiter, one that is neither adversarial to scientists nor made up of their patrons. Many research oversight regimes have been run by the NIH—the world’s largest funder of basic research—and then delegated to self-regulatory bodies such as institutional review boards and institutional biosafety committees \citep{berg1975summary,krimsky1982social}. Although this approach has the significant advantages of protecting academic autonomy and reducing political interference, it risks consigning too much authority to parties that are insufficiently objective about the risks of research they oversee \citep{bloomfield_how_grantmakers_govern_2025}. When it comes to the biosecurity risks posed by AI trained on pathogen data, end-to-end self-regulation is likely inappropriate. 

Third, the government should provide researchers with a guaranteed timeline for data classification decisions and a clear route for appealing adverse rulings. Here, the NIH’s experience with privacy controls is illuminating. The NIH generally requires its grantees to protect the privacy of individual human genetic data, and the agency has organized data-access committees, staffed by agency employees, to screen requests for genetic information. Committees can take months to consider access requests, and scientists have complained that some access decisions appear arbitrary, obscure, or directed at preventing disfavored lines of study \citep{learned2019barriers}. It is, moreover, difficult or impossible to appeal data-access decisions, just as it is challenging to appeal those of many institutional review boards. To speed valuable science and protect civil liberties, researchers should be guaranteed a fast response time on data classification and access requests, and given an opportunity to appeal government decisions before a neutral arbiter.    

Finally, the rules should apply to all scientists, not only those funded by the government. Many prior research oversight frameworks have bound only government grantees, leaving much of the private sector constrained only by voluntary commitments. For example, the NIH Recombinant and Synthetic Nucleic Acid Guidelines, the backbone of U.S. biosafety policy, applies only to institutions that accept NIH grants; the same is true of U.S. policy directed at dual-use research of concern. Significant biological model and pathogen research happens outside of government-supported institutions, and should be bound by the same biosecurity policies. 

To achieve these ends, the government should create an independent Pathogen Data Board composed of experts in artificial intelligence, biology, biosecurity, law, and ethics. Congress should task the Board with producing a set of generally applicable classification rules for governing newly produced pathogen data. The rules should provide clear, objective directives to researchers producing pathogen data. These rules should explain the types of data subject to regulation—for example, by the family of pathogens studied, the pathogen characteristics the experiments will observe, and the quantity of data collected—along with the corresponding level of controls. Table 1, along with our BDL taxonomy, represents an initial effort in this direction, but much work remains to be done to elaborate and validate regulatory tiers, and to make the data definitions objectively applicable, for example by detailing the specific viral families, data types, and data quantities that apply to each BDL tier. Because both AI and high-density pathogen data collection are advancing rapidly, the Board should revisit the details of this framework regularly. 

Running effective TREs is expensive, all the more so because the TREs in this domain must have access to secure AI-scale compute. To make it possible for small labs to comply with the Board's rules, Congress should fund private or public entities to build a compliant TRE platform, whose ongoing operations could be supported through usage fees.

To ensure compliance with the framework, principal investigators should make initial classification decisions before beginning data collection—ideally using a simple tool created by the Board—which would then be reviewed on a guaranteed timeline by institutional or regional oversight committees well-versed in the data oversight framework. The Board’s initial rules should also provide for a subset of more concerning experiment types that will be reviewed by the Board itself or a delegated subset of it. The government should give scientists and institutional committees a clear path for appealing adverse decisions to an independent body. Regulators such as the Environmental Protection Agency and the Department of Health and Human Services provide avenues for appealing decisions to independent tribunals. As with prior research oversight measures, such as laboratory safety rules, it is critical that other leading nations in biology adopt similar rules to avoid regulatory arbitrage—especially in this domain, as open data can move seamlessly across the globe. 

If Congress is unable to act quickly, agencies such as the NIH or equivalent bodies outside the United States should create initial versions of these rules themselves. There is significant precedent for science-funding agencies to step in as a gap filler before legislatures act. That was the case with human-subjects research, which the NIH pioneered at its main research hospital, the Clinical Center. It was also true when it came to biosafety. Leading scientists in genetics worked with an expert NIH panel, the Recombinant DNA Advisory Committee, to develop a set of guidelines and governing infrastructure for biosafety in the early days of genetic engineering \citep{berg1975summary}. Although data oversight rules should be generally applicable, and an NIH-only approach is undesirable, the agency can play a critical early role if Congress does not act. 

Ultimately, this framework will only be workable in the long run if all nations with significant biological research labs adopt similar rules. Although the path to international agreements in this domain will be challenging, it is no country's interest to see a diffusion of AI capabilities of concern. As with prior biological weapons threats--and as with prior large-scale data sharing undertakings like the Human Genome Project--we expect that countries can come together to agree on basic data access protocols to balance the risks and benefits of dual-use pathogen data.

\section{Conclusion}

Artificial intelligence models for biology are heavily dependent upon large volumes of relevant biological data. Some developers have attempted to remove sensitive data from training datasets in order to limit misuse-enabling capabilities, such as those for virus-related tasks. Rather than relying upon individual developers to make these safety determinations in isolation, we recommend a deliberative process to determine which biological data types may be particularly misuse-enabling, and propose an initial taxonomy here. Linked to this taxonomy, we propose technical and legal frameworks to secure sensitive data most prone to misuse. Data security and oversight requirements have historically been cumbersome, and so government and private actors must develop innovative approaches (such as data tracking tools and next-generation TREs) to limit researcher burden. Policymakers should apply restrictions narrowly, allowing the vast majority of biological data to remain easily accessible. Given the increasing accessibility of AI, and the relative costliness and expertise needed to produce biological data, governance of data may be a particularly fruitful way to reduce biological risks associated with artificial intelligence models trained on pathogen data. 

\begin{ack}
The authors thank Russell Altman, Nadav Brandes, Anita Cicero, Michael Imperiale, Thomas V. Inglesby, and Robert Pollack, with whom some of the authors have ongoing work related to BDLs, for helpful comments. The authors also thank two anonymous peer reviewers and Matt Smith for helpful feedback.
\end{ack}

\textbf{Disclosures
}
Funding (direct support): DB—Greenwall Foundation; MSH—Horizon Institute for Public Service (scholarship); AM—Open Philanthropy (Ph.D.); JRMB—Open Philanthropy; AB, TW—RAND funding from gifts from Open Philanthropy and Good Ventures; THB—NIH NCATS (1UM1TR004921), AHRQ (R01HS024096), NIH NLM (R01LM013362); OMC—New College Todd–Bird Junior Research Fellowship; MRC Fellowship MR/Y010078/1; JP—Chan Zuckerberg Initiative; Open Philanthropy.

Competing interests: THB—royalties or licenses (Coursera, AI in Healthcare); consulting fees (PAUL HARTMANN AG; Grai-Matter; Roche); stock or stock options (Verantos Inc.; Grai-Matter); advisory board (AtheloHealth). OMC—consulting fees (Pelago Biosciences; Faculty.ai; MarketCast); scientific advisory board member (Evolvere Biosciences). JP-consulting fees (Chan Zuckerberg Initiative). All others had no competing interests to declare.

No funder had a role in the research or decision to publish.

\bibliographystyle{plainnat}
\bibliography{references}





\end{document}